\documentclass[runningheads]{llncs}
\usepackage[T1]{fontenc}
\usepackage{graphicx}
\usepackage{caption}
\usepackage{amssymb}
\usepackage{amsmath}
\usepackage{dsfont}
\usepackage{booktabs} 
\usepackage{hyperref}
\usepackage{tabularx}
\usepackage{multirow}
\usepackage{makecell}
\usepackage{blindtext}
\usepackage{booktabs}
\usepackage{siunitx} 
\usepackage{array} 

\usepackage{floatrow}
\newcolumntype{Y}{>{\centering\arraybackslash}X}
\newcolumntype{C}{>{\centering\arraybackslash}X}

\newcommand{\inlinesubsection}[1]{\noindent\textbf{#1}\quad}

\begin{document}
\title{FISHing in Uncertainty: Synthetic Contrastive Learning for Genetic Aberration Detection}

\titlerunning{FISHing in Uncertainty}
%
\author{
Simon Gutwein\inst{1,2,3}\orcidID{0009-0004-8406-0736} \and
Martin Kampel\inst{2}\orcidID{0000-0002-5217-2854} \and
Sabine Taschner-Mandl\inst{1}\orcidID{0000-0002-1439-5301}
\and Roxane Licandro\inst{3}\orcidID{0000-0001-9066-4473}
}
\authorrunning{Gutwein et. al}
%
\institute{St. Anna Children's Cancer Research Institute, Vienna, Austria \and TU Wien, Institute of Visual Computing and Human-Centered Technology, Computer Vision Lab, Vienna, Austria \and Medical University of Vienna, Department of Biomedical Imaging and Image-guided Therapy, Computational Imaging Research, ELIA Group, Vienna, Austria}
%
\maketitle              
\begin{abstract}

Detecting genetic aberrations is crucial in cancer diagnosis, typically through fluorescence \textit{in situ} hybridization (FISH). However, existing FISH image classification methods face challenges due to signal variability, the need for costly manual annotations and fail to adequately address the intrinsic uncertainty. We introduce a novel approach that leverages synthetic images to eliminate the requirement for manual annotations and utilizes a joint contrastive and classification objective for training to account for inter-class variation effectively. We demonstrate the superior generalization capabilities and uncertainty calibration of our method, which is trained on synthetic data, by testing it on a manually annotated dataset of real-world FISH images. Our model offers superior calibration in terms of classification accuracy and uncertainty quantification with a classification accuracy of 96.7\% among the 50\% most certain cases. The presented end-to-end method reduces the demands on personnel and time and improves the diagnostic workflow due to its accuracy and adaptability. All code and data is publicly accessible at: \href{https://github.com/SimonBon/FISHing}{https://github.com/SimonBon/FISHing}.


\end{abstract}
\section{Introduction}
\begin{figure}[!t]
    \centering
    \includegraphics[width=\textwidth]{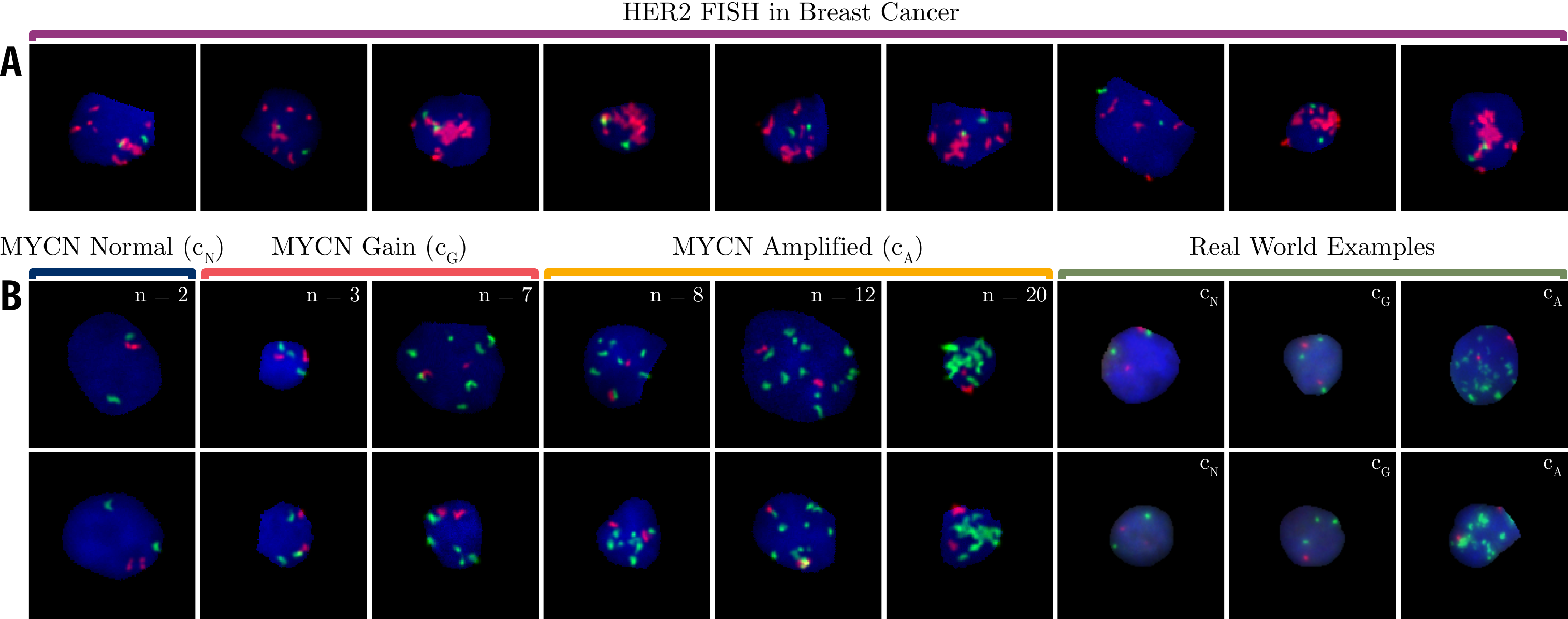}
    \caption{FISH-image patches with stained cell nuclei in blue image channel for row \textbf{A} and \textbf{B}. \textbf{A}:~\textit{HER2} (\textit{target}) as red signals, \textit{CEN17q} (\textit{reference}) as green signals. \textbf{B}: Synthetic images of \textit{MYCN} FISH, \textit{MYCN} (\textit{target}) as green signals, \textit{NMI} (\textit{reference}) as red signals, showing diagnostic classes: \textit{MYCN Normal} ($c_N$), \textit{Gain} ($c_G$), and \textit{Amplified} ($c_A$) along real world image examples. \textit{n} indicates the number of \textit{MYCN} signals.}
    \label{fig:example_patches}
\end{figure}

Evaluating genetic aberrations is crucial in cancer diagnostics, including breast \cite{Penault2009}, lung cancer \cite{Tang2019} or neuroblastoma \cite{Ambros2009}. Fluorescence \textit{in situ} hybridization (FISH) followed by fluorescence microscopy, allows for the visualization of gene copies in cell nuclei. Fluorophore-labeled probes are employed to bind to DNA sequences of a gene of interest (\textit{target}) and a \textit{reference} gene and are imaged using distinct wavelengths. FISH signals manifest as red and green spots in an RGB color scheme, with the nucleus stained blue as illustrated in Fig.~\ref{fig:example_patches}A for \textit{HER2} in breast cancer and Fig.~\ref{fig:example_patches}B for \textit{MYCN} in neuroblastoma risk stratification \cite{Cohn2009}.
\vspace{4pt}

\inlinesubsection{Challenges:} Assessing gene copy number in FISH images requires expert manual evaluation to count the signals.
It is a tedious and subjective process, which embodies inherent uncertainty, particularly when signal proximity or cluster formation complicates the precision of gene copy number determinations (see Fig.~\ref{fig:example_patches}A).
This necessitates multiple expert reviews, leading to increasing costs and time. 

\vspace{4pt}
\inlinesubsection{Related Work:} 
Current FISH image analysis approaches, including machine and deep learning\cite{Gudla2017,Eichenberger2021,Wollmann2019,Bouilhol2021}, aim at direct spot identification or appearance standardization for easier application of traditional methods like thresholding or gradient techniques \cite{Bahry2021,Imbert2022,Tinevez2017}. 
While certain methods excel with images having clear spot-like signals, they falter with FISH image variability, especially in case of gene amplifications lacking defined spot appearances (see Fig.~\ref{fig:example_patches}B n=20). 
In \cite{Zakrzewski2019} a \textit{RetinaNet}-based dual-network approach has been devised for \textit{HER2} status identification in breast cancer. It classifies nuclei and signals independently, but inaccurately estimates gene copy counts in clusters, defaulting to a fixed number - not necessarily mirroring reality - and inadequately addresses prediction uncertainty. 
In \cite{Gutwein2024} a two-stream model is introduced by combining whole image and spot information, for FISH image classification, requiring minimal manual annotations, but raising questions about detection reliability and lack of uncertainty quantification.
In \cite{Imbert2023} point clouds were applied to represent localization patterns in synthetic single molecule FISH images effectively, yet this method demands pre-extraction of point positions, limiting its direct applicability to imaging data.
To date, no studies have successfully integrated uncertainty modeling into FISH classification.

\vspace{4pt}
\inlinesubsection{Contribution:} Our work provides four major contributions: 
(i) An innovative methodology for the precise classification of single-cell FISH images, leveraging synthetic data to bypass the need for costly manual annotations. 
(ii) A technique for generating the aforementioned synthetic FISH images. 
(iii) An innovative strategy for embedding inter-class variation into classification through a contrastive learning-based joint objective. 
(iv) Our approach enables the accurate quantification of classification uncertainty, demonstrating its concordance with human expert judgment and its applicability in the diagnostic processes.
\section{Material \& Methods}
The proposed work consists of two modules. First, we introduce "FISHPainter" a method offering users the flexibility to create FISH images with desired signal characteristics and second, a novel contrastive learning (CL) approach for genetic aberration classification incorporating the uncertainty of classification without manual annotations.
\vspace{4pt}%

\inlinesubsection{FISHPainter:} To address the data sparsity in the field of FISH imaging, especially given its extensive applications across various gene targets and diagnostic use cases, we propose FISHPainter. This tool generates synthetic images, giving users complete control over the modeled FISH modality and data distribution, thereby creating a diverse and well-populated image space, which real-world images, with their prevalence biases, are not sure to provide. Importantly, FISHPainter enables the population of image spaces for cases that lie on the boundaries of different diagnostic categories, which in turn is expected to improve uncertainty estimation in these critical areas. \\ FISHPainter involves the following steps:
(i) Selecting a cell nucleus background from a predefined library to place at the center of an RGB image patch in the blue channel. (ii) Defining positions for signals in the red and green channel, distinguishing between individual and clustered signals — the former randomly placed within the nucleus, and the latter formed by adding signals near a central point, all represented as 2D-Gaussian distributions with user-defined sizes. 3) Applying non-affine transformations to signals to vary appearances. Fig.~\ref{fig:example_patches}B illustrates generated patches, with $n$ indicating the number of green signals. Implementation available at: \href{https://github.com/SimonBon/FISHPainter}{https://github.com/SimonBon/FISHPainter}
\vspace{4pt}

\inlinesubsection{Synthetic Contrastive Learning:} Studies show that classification methods based solely on Cross Entropy (CE) inaccurately gauge uncertainty, leading to overconfidence \cite{Wei2022,Algan2021}. Alternative techniques, including Ensembles \cite{Lakshminarayanan2016}, Monte-Carlo Dropout \cite{Gal2015}, and Multi-Head models \cite{Linmans2020}, aim to enhance uncertainty estimation, though they face scalability issues due to their high computational requirements. CL has been recognized for its capacity to incorporate uncertainty into data representations \cite{Wu2020,Ardeshir2022,Kirchhof2023}.
Our method, inspired by \cite{Winkens2020,Mukhoti2023}, infers prediction uncertainty from softmax outputs of high-quality latent representations of FISH images, jointly training the network's backbone and classification head using CL.
Acknowledging that class label-based representations often miss class variability and therefore inherent uncertainty, our model integrates both class labels and visual similarity into its latent representation, using CE and CL loss. 
This approach accounts for the levels of intra-class variation by combining NT-Xent \cite{Sohn2016} and CE in the loss function:   

\begin{equation}
\scriptsize
\mathcal{L} = - \left(  \underbrace{\log \frac{\exp(\text{sim}(z_i, z_j)/\tau)}{\sum_{k=1}^{2N} \mathds{1}_{[k \neq i]} \exp(\text{sim}(z_i, z_k)/\tau)}}_{\text{Contrastive}} + \underbrace{\lambda \left( \sum_{c=1}^{C} y_c \log(\hat{y}_{i,c}) + \sum_{c=1}^{C} y_c \log(\hat{y}_{j,c}) \right)}_{\text{CE for Positive Pair $(i, j)$}} \right)
\label{eq:loss_function}
\end{equation}

\noindent In Eq.~\ref{eq:loss_function}, $N$ is the batch size, $z_i$ and  $z_j$ are the projected embeddings of two augmented views $i$ and $j$ of a patch, $\tau$ is the temperature parameter, and $\lambda$ is the classification loss weight. $C$ is the class count, $y_c$ and $\hat{y}_{\cdot,c}$ are the target and predicted probabilities for class $c$, respectively and $sim$ the similarity metric . 
\vspace{4pt}

\begin{figure}[b!]
    \centering
    \includegraphics[width=0.8\textwidth]{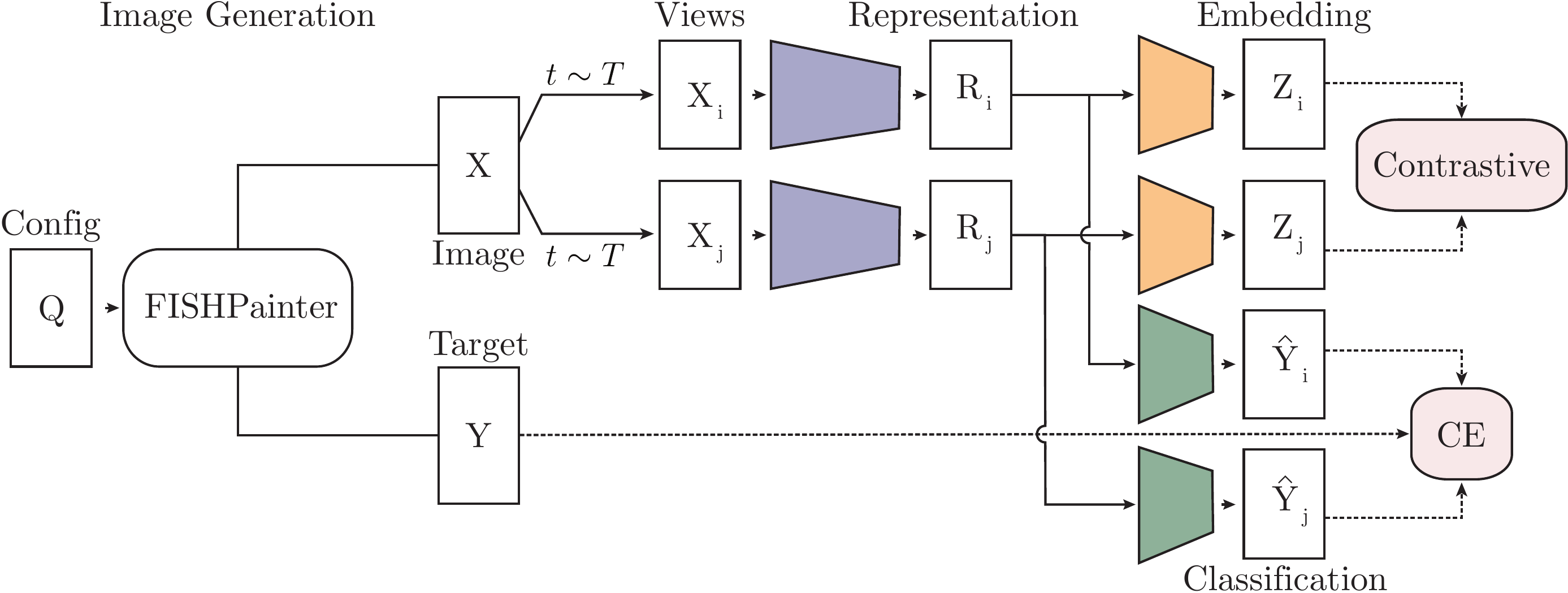}
    \caption{Configuration of signals $Q$ is input to FISHPainter, producing image $ X $ and label $ Y $. Image $ X $ is augmented with transformations $t$ sampled from set $ T $ to create views $ X_i $ and $ X_j $. These views are embedded into representation space $ R $ and projected to $ Z $. Contrastive loss is calculated on $ Z $. $ R $ is also used by the classification layer to predict $ \hat{Y} $, on which CE loss is computed.}
    \label{fig:workflow}
\end{figure}

\noindent In Fig.~\ref{fig:workflow} the model training using the loss function introduced in Eq.~\ref{eq:loss_function} is shown: First, for generating input images $X$ and target labels $Y$ with FISHPainter, a set of user-defined signal configurations $Q$ are utilized. 
The resulting images are augmented with transformations $t$ sampled from a set of transformations $T$, producing views $X_i$ and $X_j$. These views are embedded by the network's backbone (blue) into the representation space $R$. This representation is fed into the projector (orange) to obtain an embedding $Z$, on which the contrastive loss is calculated, and into the classification head (green), producing a class prediction $\hat{Y}$ for CE loss computation using the target label $Y$.
\vspace{4pt}

\inlinesubsection{Uncertainty:} 
Uncertainty estimation is crucial in medical artificial intelligence applications \cite{Seoni2023}. Epistemic uncertainty helps to identify out-of-distribution (OOD) samples \cite{Winkens2020}, while aleatoric uncertainty addresses inherent data uncertainties, such as ambiguous signal quality. In this work our main focus lies on aleatoric uncertainty, which we will refer to in the following as uncertainty. Inspired by \cite{Mukhoti2023}, we model aleatoric uncertainty $\mathbb{H}_{\text{norm}}(.)$ using normed entropy using  Eq. \ref{eq:H}:

\begin{equation}\label{eq:H}
\mathbb{H}_{\text{norm}}(\hat{Y}) = \frac{\mathbb{H}(\hat{Y}) - \mathbb{H}_{\text{min}}\left(\alpha, C\right) }{\log(C)}
\end{equation}

\noindent Where $\mathbb{H}(\cdot)$ denotes the entropy function, $\hat{Y}$ represents the softmax output of the model, $C$ is the number of classes, and $\mathbb{H}_{\text{min}}(\alpha, C)$ indicates the minimum entropy for label smoothing value $\alpha$, which is calculated as follows: \begin{equation}\label{eq:Hmin}
\mathbb{H}_{\text{min}}\left(\alpha, C\right) = \mathbb{H}\left(\left[ 1-\alpha, \left(\frac{\alpha}{C-1}\right)_0, \left(\frac{\alpha}{C-1}\right)_1, \ldots, \left(\frac{\alpha}{C-1}\right)_{C-1} \right]\right)
\end{equation}
\section{Experimental Setup}\label{sec:expSetup}
The performance of the proposed method is evaluated on the task of classifying \textit{MYCN} gene aberrations in neuroblastoma \cite{Ambros2009} with the following three classes: \textit{MYCN Normal} ($c_N$) - showing 2 green and 2 red signals (2:2 - \textit{MYCN}:\textit{NMI} ratio), \textit{MYCN Gain} ($c_G$)  - showing 3-7 (green) and 2 (red) and \textit{MYCN Amplification}~($c_A$) with $\geq8$ (green) and 2 (red), as shown in Fig.~\ref{fig:example_patches}B. We therefore set $C=3$ in Eq.~\ref{eq:H} and Eq.~\ref{eq:Hmin}.
\vspace{4pt}

\inlinesubsection{Data Setup 1:} In experiment 1 the generalisation from synthetic FISH data training to real world FISH image classification is analysed using a training dataset of 30,000 synthetic images (10,000 each of $c_N$, $c_G$ and $c_A$) utilizing "FISHPainter" (see configurations $Q$ in Supplementary Fig.~\ref{fig:augmentations}A). We used a 60\% training, 20\% validation, and 20\% testing split. Additionally, we employ an expert annotated real-world single-cell \textit{MYCN} FISH dataset containing 1814 single-nuclei image patches, obtained from 2 $c_N$, 2 $c_G$ and 2 $c_A$ cell lines, and use it to assess model generalization in experiment 1, and uncertainty calibration in experiment 3.
The following set of augmentations $T$ were chosen to model variations in microscopic images: rotations, flipping, scaling, blurring, channel-specific intensity adjustments, noise, and gradient additions (see parameters and examples in Supplementary Fig.~\ref{fig:augmentations}B).
\vspace{4pt}

\inlinesubsection{Data Setup 2:} In experiment 2 the aim is to assess the method's ability to detect in distribution (ID) and OOD samples. The synthetic training dataset serves as the reference, while the 1814 expert annotated real-world FISH images are considered in-distribution (\textit{Real World ID}).
For OOD analysis we use an in-house immunofluorescence dataset (\textit{OOD}), similar yet distinct from FISH (see Fig.~\ref{fig:examples_uncertain}C) in targeting cell surface proteins instead of genetic sequences.
\vspace{4pt}

\inlinesubsection{Data Setup 3:} In experiment 3, we focused on assessing how well the model matches human uncertainty by utilizing annotations of ten experts on a set of 1210 synthetic single-cell FISH images with predefined number of green signals. Images were annotated either $c_N$, $c_G$ or $c_A$. Accuracy was measured across all images, while human uncertainty was calculated per image using normalized entropy on the expert annotations.
\vspace{4pt}

\inlinesubsection{Method Implementation:} We implemented our approach utilizing two sets of augmentations $T$ having different intensity levels: \textit{Ours Heavy}, which applies stronger augmentations, and \textit{Ours Light}, with lighter augmentations (details see Supplementary Fig.~\ref{fig:augmentations}B). Both utilize the following parameters: A ResNet-18 \cite{He2015} backbone modified to reduce the representation space dimension to 128 with a two-layer 64-dimension projection head for CL training \cite{Chen2020}. Classification relied on 2 fully connected layers with 128 features, 0.25 dropout, and ReLU activation. The loss function included cosine similarity, 0.05 temperature, 0.5 as $\lambda$, and 0.01 as label smoothing ($\alpha$). We used a Cosine Annealing learning rate scheduler with 5 warmup steps and a 25-epoch cycle, setting learning rate minimum and maximum at 0.00001 and 0.001 respectively, and a batch size ($N$) of 128.
\vspace{4pt} 

\inlinesubsection{Baselines:} We compared our method against \textbf{bigFISH} \cite{Imbert2022} and a pretrained spot detection \textbf{RetinaNet} \cite{Lin2017} model from \cite{Gutwein2024}, with a classification head fine-tuned on our synthetic training datasets. Additionally, we used three baselines, all employing a \textbf{ResNet-18} backbone: (i) \textbf{ResNet+CE}: trained using only CE loss, and two baselines using CL with NT-Xent loss and cosine similarity \cite{Sohn2016}, (ii) \textbf{CL+Detached}: pretrained with contrastive loss, then froze the backbone and fine-tuned only the classification head with CE, (iii) \textbf{CL+Attached}: pretrained with contrastive loss, then fine-tuned both the backbone and the classification head with CE. The set of augmentations \( T \) for all baselines can be seen in Fig.~\ref{fig:augmentations}B. Fig.~\ref{fig:all_others} shows a schematic illustration of each baseline's training schemes.
\section{Results}
\inlinesubsection{Experiment 1:} The ablation study assesses augmentation effectiveness using classification accuracy across the synthetic test split and the real-world data (Data Setup 1). We compared the use of all augmentations to no augmentations and employed a leave-one-out approach, omitting one augmentation at a time (see Table \ref{tab:augmentations}). Augmentations have a minor effect on test split accuracy—94.3\% without augmentations versus 94.6\% with them. Yet, for real-world datasets, augmentations significantly boost generalization, increasing accuracy from 40.6\% to 87.8\% without affine transformations, underscoring their critical role in bridging the synthetic and real-world accuracy gap. Gradient and noise augmentations show the greatest impact when ommited (see Table \ref{tab:augmentations} column Grad\&Noise).
\vspace{4pt}

%
\begin{table}
\centering
\caption{Classification accuracies (in \%) for the synthetic test set and the manually annotated real-world dataset. Int.: intensity scaling, Grad.: image gradient.}
\begin{tabularx}{\textwidth}{lYY|YYcYcYc}

& \multicolumn{2}{c}{Augmentations} & \multicolumn{7}{c}{Augmentations Omitted}\\
\toprule
& None  & All  & Affine & Blur & Flip & Grad. & Noise & Int. & Grad\&Noise\\
\midrule
Synth. Test Set   & 94.3  & 94.6  & 94.7   & 95.1 & 95.2 & 94.7 & 95.2  & \textbf{96.0} & 94.9\\
Real World & 40.6 & 86.1 & \textbf{87.8}   & 85.4 & 87.2 & 80.5  & 80.8  & 86.7 & 45.0\\
\bottomrule
\end{tabularx}
\label{tab:augmentations}
\end{table}

\begin{figure}[h!]
    \centering
    \includegraphics[width=0.95\textwidth]{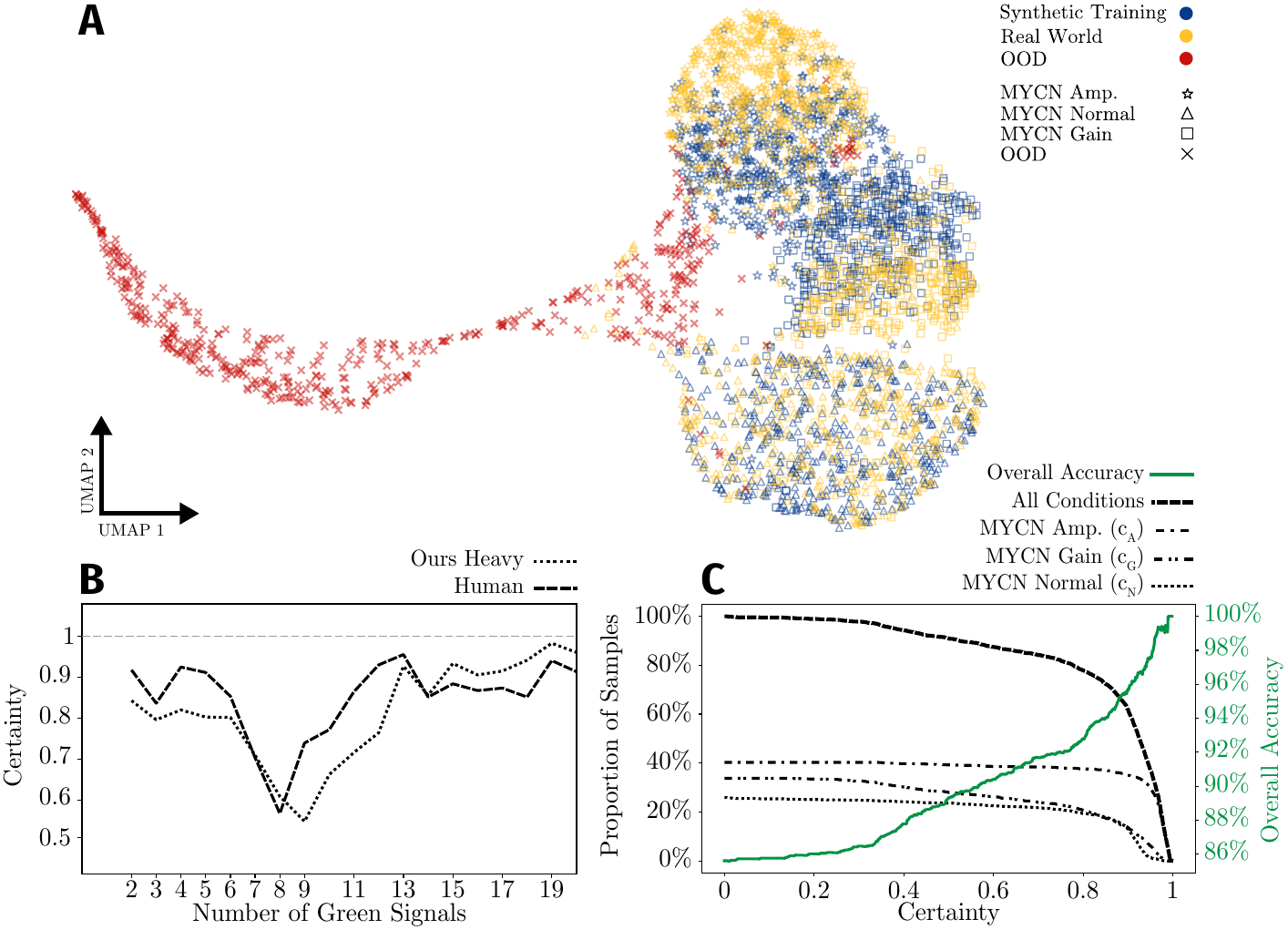}
    \caption{\textbf{A}: 2D latent space visualization of \textit{Ours Heavy} on the synthetic training dataset (blue), the real world test dataset (yellow) and the OOD dataset (red). \textbf{B}: Comparison of model to human certainty when classifying synthetic FISH images with set number of green signals. \textbf{C}: Accuracy results and the distribution of remaining classes, conditioned on certainty (example: at 0.8 certainty $\sim78\%$ data remaining with $\sim93\%$ overall accuracy).}
    \label{fig:epistemic}
\end{figure}

\inlinesubsection{Experiment 2:} The detection of OOD and ID samples is visually evaluated using Data Setup 2. 
Fig.~\ref{fig:epistemic}A shows a UMAP~\cite{McInnes2020} of the final latent representation of \textit{Ours Heavy} (see Fig.~\ref{fig:all_others} for other baselines). The real-world dataset significantly overlaps with the synthetic training dataset, showing strong generalization, while the majority of OOD samples occupy a distinct area of the feature space, making them easily separable from ID data.
Fig.~\ref{fig:examples_uncertain}C shows OOD samples embedded in an ID latent region, illustrating signals that could be mistaken for clustered \textit{MYCN} signals.
\vspace{4pt}

\inlinesubsection{Experiment 3:} Accuracy of ten human annotators and our model on Data Setup 1 is presented in Table~\ref{tab:humanvsai}. Our method achieves 90.5\% accuracy, close to the top annotator's 90.6\% and exceeds the expert average by 2\%. Fig.~\ref{fig:epistemic}B shows the alignment between model-generated aleatoric uncertainty, calculated following Eq.~\ref{eq:H} and inter-expert agreement, highlighting similar uncertainty patterns between our model and the annotators, especially around 8 green signals, which is the classification border between $c_A$ and $c_G$.\\ 

\begin{table}
\centering
\caption{Comparison of \textit{Ours Heavy} and human annotator accuracy}
\begin{tabularx}{\textwidth}{cccccccccccCc}
\toprule
Annotator & 1 & 2 & 3 & 4 & 5 & 6 & 7 & 8 & 9 & 10 & \makecell{Combined \\ mean ± std} &  \makecell{Ours \\ Heavy} \\
\midrule
Accuracy [\%] & 89.3 & 90.3 & 88.0 & 84.7 & 88.3 & 89.0 & 89.0 & 88.9 & 86.9 & \textbf{90.6} & 88.5 ± 1.6 & 90.5 \\
\bottomrule
\end{tabularx}
\label{tab:humanvsai}
\end{table}

\begin{figure}[!t]
    \centering
    \includegraphics[width=0.9\textwidth]{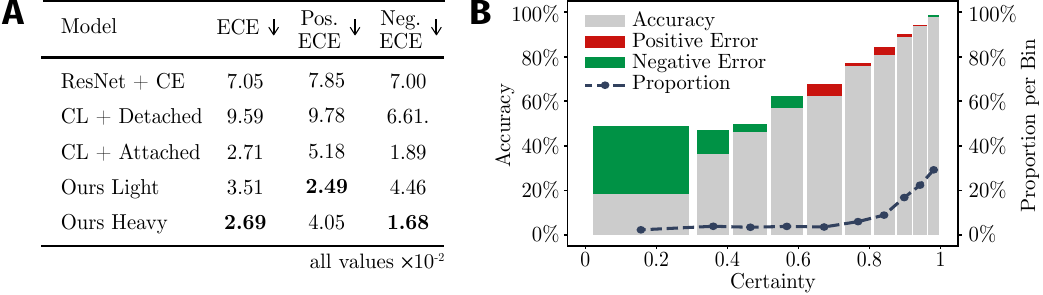}
    \caption{\textbf{A}: ECE, positive ECE, and negative ECE comparison between our method and all baseline approaches on real world data. \textbf{B}: Calibration chart for \textit{Ours Heavy}, showing the sample distribution across certainty bins.}    
    \label{fig:ECE_table}
\end{figure}

\noindent We assess the Expected Calibration Error (ECE) \cite{Ding2019} including positive ECE (overconfidence) and negative ECE (underconfidence). Fig.~\ref{fig:ECE_table}A shows our approach's ECE results using annotated real-world data from Data Setup 1 compared to the baselines. 
Fig.~\ref{fig:ECE_table}B presents the calibration diagram, featuring a curve indicating data distribution across certainty levels for \textit{Ours Heavy}.
The calibration diagram for baseline methods is visualized in Fig.~\ref{fig:all_others}.
Our method is well calibrated with the highest overall and negative ECE using heavy augmentations and the lowest positive ECE using light augmentations.
In Fig.~\ref{fig:examples_uncertain} instances of high model uncertainty are presented highlighting ambiguity, particularly when classifying  $c_N$ and $c_G$ (A), where distinguishing between 2 or 3 \textit{MYCN} signals is subjective, and between $c_G$ and $c_A$ (B), where determining the exact \textit{MYCN} copy number is challenging. This affirms that model-generated- and human- uncertainty align, given that these instances are inherently ambiguous and their classification ultimately relies on subjective judgment.\\
Fig. \ref{fig:epistemic}C shows the accuracy and class distribution for \textit{Ours Heavy} conditioned on certainty thresholds, excluding samples below the threshold. Initially, uncertain samples from $c_G$ are removed, followed by uncertain $c_N$ samples, and finally $c_A$ cases. This indicates that $c_A$ cases are easier to classify due to their distinct visual differences from $c_G$ or $c_N$ (compare n=20 with n=2 and n=5 in Fig. \ref{fig:example_patches}B). Conditioning plots for all baselines are shown in Fig.~\ref{fig:all_others}.\\ ~
Table \ref{tab:conditioning} shows accuracy scores for each baseline at different remaining sample percentages, conditioned on the yielded aleatoric certainty, calculated following Eq.~\ref{eq:H}. Since \textit{bigFISH} and \textit{RetinaNet} do not provide uncertainties, their accuracy is measured only at 100\%. Accuracy values indicate they cannot compete with models that do not rely on spot signals. The \textit{CL+Attached} model performs well with 88.3\% accuracy at 100\% data but becomes unstable for higher certainty values (see 20\% - 5\%). Notably, \textit{Ours Light} shows 90.3\% accuracy at 95\% data retention, and \textit{Ours Heavy} achieves 98.9\% accuracy at 30\% of the most certain samples, further increasing to 100\% at 5\%. \textit{ResNet+CE} exhibits overconfidence, reaching maximum certainty for the top 40\% of samples but only achieving 95.6\% accuracy, highlighting its overconfidence, which is also observable in ECE plot of Fig.~\ref{fig:all_others}A. In contrast, \textit{Ours Heavy} improves accuracy as the conditioning becomes more stringent. We hypothesize that the performance difference between \textit{Ours Heavy} and \textit{Ours Light} is due to a more restricted feature space under the contrastive loss in \textit{Ours Heavy}. This leads to poorer cluster separation in the embedding space, but better calibration, particularly for positive ECE and overconfidence. Additionally, we observe that lighter augmentation intensity improves class cluster separation in the embedding space, but this results in poorer calibration by not adequately representing the transition between classes.

\begin{table}[!h]
\caption{Accuracy comparisons between our method and baseline models after excluding the most uncertain samples, based on conditioning on certainty. "-" indicates that all data has been removed, so accuracy can not be calculated.}
\centering
\begin{tabularx}{\textwidth}{Xcccccccccccc} 
\toprule
                         & & \multicolumn{11}{c}{Percentage of remaining data}                            \\
Model Type              &  & 100\% & 95\% & 90\% & 75\% & 50\% & 40\% & 30\% & 20\% & 15\% & 10\% & 5\%   \\ 
\midrule
bigFISH \cite{Imbert2022} & \multirow{7}{*}{\rotatebox{90}{Accuracies [\%]}\hspace{5pt}}  & 59,4  &   &  &  & \multicolumn{4}{c}{not applicable} & &  &   \\
RetinaNet   \cite{Gutwein2024} && 71,9   &   &  &  & \multicolumn{4}{c}{not applicable} &  &  &    \\
ResNet+CE           &   & 86,4  & 88,5 & 90,8 & 94,7 & 95,6 & 95,6 & - & - & - & - & -  \\
CL+Detached               &   & 73    & 73,8 & 75,3 & 80,4 & 89   & 93,3 & 96,2 & 98,1 & 97,8 & 97,2 & 98,9  \\
CL+Attached               &   & \textbf{88,3}  & 90,1 & 91,4 & 94,1 & 96,3 & 96,7 & 96,7 & 95,6 & 94,7 & 92,7 & 93,1  \\
Ours Light               &   & \textbf{88,3}  & \textbf{90,3} & \textbf{91,7} & \textbf{95,2} & 96,0 & 96,4 & 96,2 & 96,9 & 98,2 & 97,9 & 100   \\
Ours Heavy               &   & 85,6  & 87,4 & 89,6 & 93,8 & \textbf{96,7} & \textbf{97,5} & \textbf{98,9} & \textbf{99,2} & \textbf{99,3} & \textbf{99,4} & \textbf{100}   \\
\bottomrule
\end{tabularx}
\label{tab:conditioning}
\end{table}
\section{Conclusion} %
We introduced an adaptable, end-to-end methodology for classifying FISH single-cell images without manual annotations using "FISHPainter", which only requires a clear class definition. A key aspect of our approach is the use of synthetic data for training, enabling us to densely model the image space of FISH images. Our results demonstrate that FISH-specific augmentations are essential for transitioning from synthetic training to real-world test settings. By using synthetic FISH images with appropriate augmentations and incorporating cross-entropy into the contrastive loss, our method produces a well-calibrated model. This integration of uncertainty is crucial for advancing digital pathology diagnostics, providing more reliable and interpretable results.\\

\begin{credits}
 \subsubsection{\ackname} 
This research was supported by Vienna Science and Technology Fund (WWTF) PREDICTOME [10.47379/LS20065], EU EUCAIM (No.101100633-EUCAIM) and the Austrian Science Fund (FWF) MAPMET [10.55776/P35841].
\end{credits}
%
%
%
\bibliographystyle{splncs04}
\bibliography{LaTeX-UNSURE/main.bib}

\newpage

\section{Supplementary Material}

\setcounter{figure}{0} 
\renewcommand{\thefigure}{S\arabic{figure}} 
\setcounter{table}{0} 
\renewcommand{\thetable}{S\arabic{table}} 

\begin{figure}[!h]
    \centering
    \includegraphics[width=0.99\textwidth]{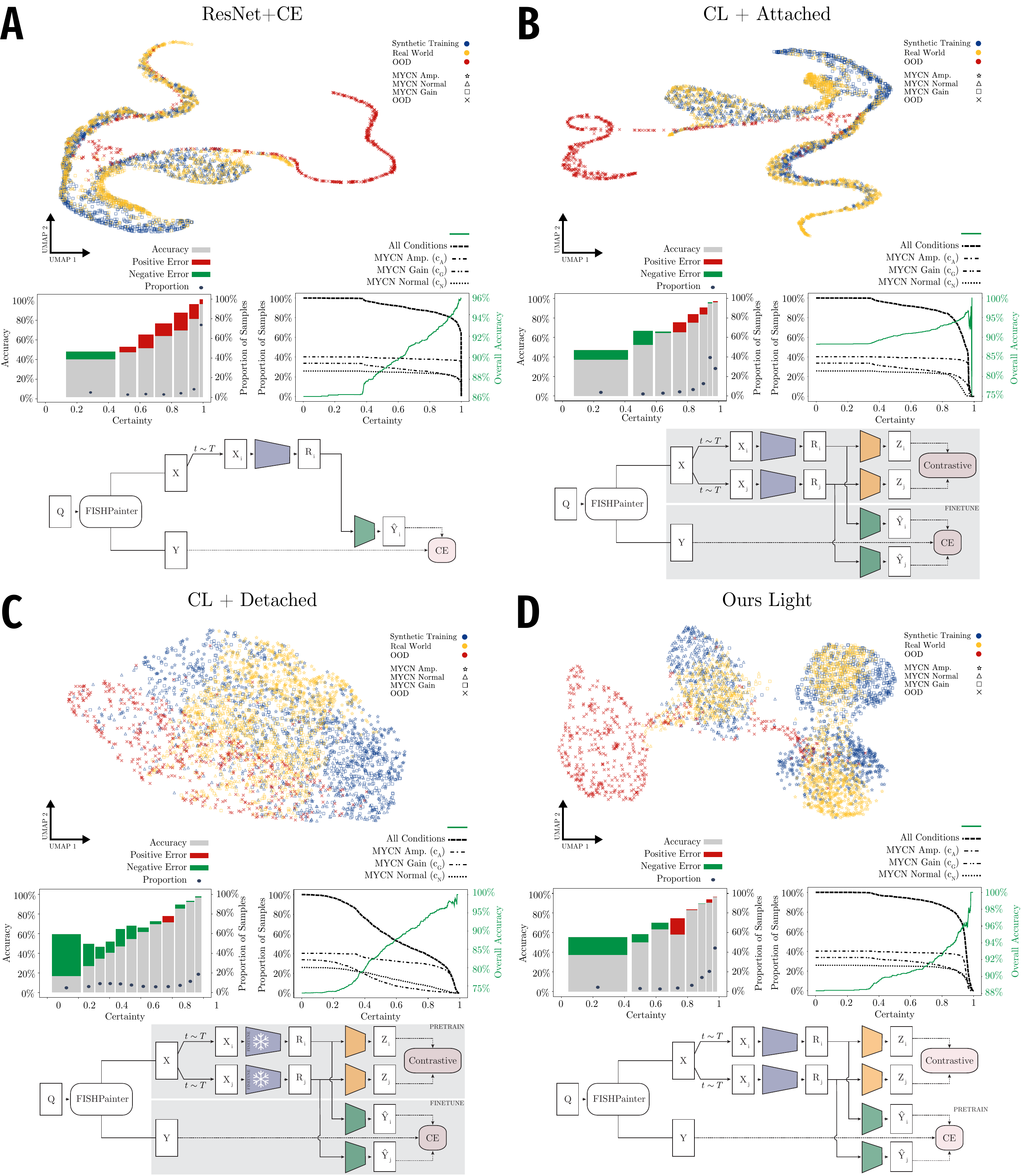}
    \caption{The experimental results are shown in subplots \textbf{A-D} for Resnet+CE, CL+Attached, CL+Detached, and Ours Light. Each subplot displays latent representations of the synthetic training dataset (blue), real-world test set (yellow), and OOD dataset (red). Below, the left plot shows ECE (green for underconfidence, red for overconfidence), and the right plot shows accuracy and class distribution across certainty thresholds. A data flow diagram below these plots illustrates the training methods for each model. The snowflake in C indicates frozen weights during fine-tuning.}
    \label{fig:all_others}
\end{figure}

\begin{figure}[!h]
    \centering
    \includegraphics[width=\textwidth]{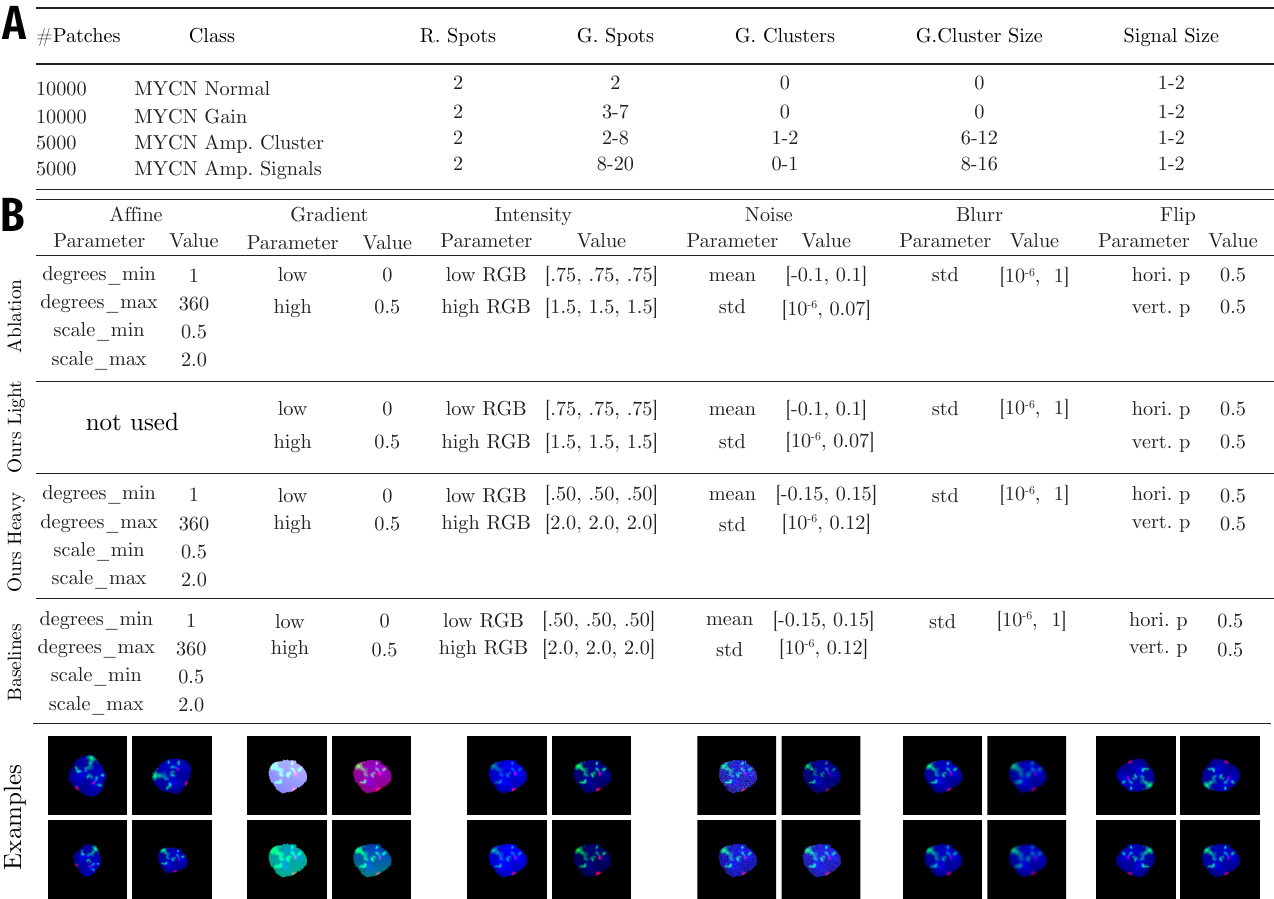}
    \caption{\textbf{A}: Configurations $Q$ of synthetic dataset: Classes are \textit{MYCN Normal}, \textit{MYCN Gain}, and \textit{MYCN Amplified}, with details on spot counts, cluster presence, and signal sizes to simulate diverse FISH image scenarios. \textit{MYCN Amplified} is divided into "Cluster" and "Signals". R.: red, G.: green. \textbf{B}: Specification of augmentations for the ablation experiment and the two implementations of our approach, \textit{Ours Heavy} and \textit{Ours Light}, along with examples showing the effect of the augmentations on the same cell.}
    \label{fig:augmentations}
\end{figure}

\begin{figure}[!h]
    \centering
    \includegraphics[width=\textwidth]{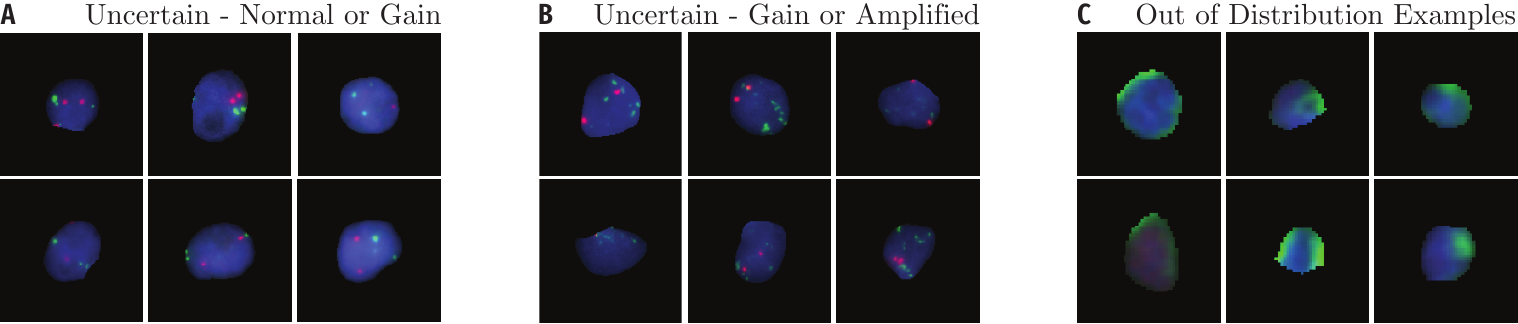}
    \caption{\textbf{A-B}: High Uncertainty Cases for \textit{Ours Heavy}. \textbf{A}: Images challenging for both humans and Ours Heavy to classify as \textit{MYCN Normal} or \textit{MYCN Gain}. \textbf{B}: Examples with ambiguous \textit{MYCN} copy numbers, hard to classify as \textit{MYCN Gain} or \textit{MYCN Amplified}. \textbf{C}: OOD images embedded close to ID samples.}
    \label{fig:examples_uncertain}
\end{figure}

\end{document}